\documentclass[sigconf]{acmart}

\usepackage{amsmath,amsfonts}
\usepackage{algorithmic}
\usepackage{graphicx}
\usepackage{textcomp}
\usepackage{bmpsize}
\usepackage{balance}
\usepackage{xcolor}
\usepackage{tabularx}
\usepackage{multirow, makecell}

\graphicspath{{figs/}}

\AtBeginDocument{%
  \providecommand\BibTeX{{%
    \normalfont B\kern-0.5em{\scshape i\kern-0.25em b}\kern-0.8em\TeX}}}
    
\renewcommand\footnotetextcopyrightpermission[1]{} 
\setcopyright{none}

\acmDOI{}
\acmISBN{}

\acmYear{2024} 
\acmConference[MiLeTS '24]{}{August 25, 2024}{Barcelona, Spain}
\acmPrice{}
       
\begin{document}
\title[Towards Foundation Auto-Encoders for Time-Series Anomaly Detection]{Towards Foundation Auto-Encoders for Time-Series\\Anomaly Detection}

\author{Gast\'on Garc\'ia Gonz\'alez, ($\dagger$) Pedro Casas, Emilio Mart\'inez, Alicia Fern\'andez}
\affiliation{%
  \institution{IIE-FING, Universidad de la Rep\'ublica; ($\dagger$) AIT Austrian Institute of Technology}
  \city{gastong@fing.edu.uy, pedro.casas@ait.ac.at, emartinez@fing.edu.uy, alicia@fing.edu.uy}
  \country{}
}

\renewcommand{\shortauthors}{Garc\'ia Gonz\'alez et al.}

\begin{abstract}
We investigate a novel approach to time-series modeling, inspired by the successes of large pretrained foundation models. We introduce FAE (Foundation Auto-Encoders), a foundation generative-AI model for anomaly detection in time-series data, based on Variational Auto-Encoders (VAEs). By foundation, we mean a model pretrained on massive amounts of time-series data which can learn complex temporal patterns useful for accurate modeling, forecasting, and detection of anomalies on previously unseen datasets. FAE leverages VAEs and Dilated Convolutional Neural Networks (DCNNs) to build a generic model for univariate time-series modeling, which could eventually perform properly in out-of-the-box, zero-shot anomaly detection applications. We introduce the main concepts of FAE, and present preliminary results in different multi-dimensional time-series datasets from various domains, including a real dataset from an operational mobile ISP, and the well known KDD 2021 Anomaly Detection dataset.
\end{abstract}


\maketitle

\section{Introduction}\label{sec:introduction}

While the literature offers a plethora of traditional statistical models for anomaly detection in time-series data, they often struggle with the non-stationary, non-linear, and noisy nature of real data, leading to suboptimal predictions. In recent years, there has been a surge in the adoption of modern deep learning-based approaches for time-series anomaly detection \cite{Guansong2021, adts2022}, owing to their ability to handle complex dependencies and generate realistic data sequences. Generative AI methodologies, in particular, have gained attention for their performance in time-series modeling \cite{lai2018stochastic, timeGAN2019, ggg2021generative}.

In this paper, we focus on devising a Generative AI model capable of matching or even surpassing the performance of conventional time-series modeling methods without the need for training on the specific target dataset - a concept known as Zero-Shot Learning (ZSL). ZSL is a problem setup in deep learning where, at test time, a learner observes samples from classes which were not observed during training, and needs to predict the class that they belong to. We therefore investigate if a model pretrained on multiple time-series data can learn temporal patterns useful for accurate forecasting on previously unseen time-series. For doing so, we use as starting point the DC-VAE model \cite{dcvae_ieee_2023}, a deep-learning-based, unsupervised, and multivariate approach to real-time anomaly detection in MTS data, based on popular Variational Auto-Encoders (VAEs) \cite{kingma2013auto}. 

VAEs are generative AI models that learn the underlying distribution of the data and can generate new samples from this distribution. In the context of time-series data, VAEs can capture latent representations of temporal patterns and generate sequences that exhibit similar characteristics, making them powerful for generalization and ZSL. VAEs learn a low-dimensional latent space representation of the input data, which captures the underlying structure of the data in a compressed form. By learning meaningful representations, VAEs can generalize well to unseen data points that lie within the same distribution as the training data, supporting generalization to new instances. As generative models, VAEs can generate new samples from the learned latent space distribution. potentially enabling ZSL, as the model can produce samples that belong to unseen classes or categories without explicitly training on them. By sampling from the latent space, VAEs can generate diverse and realistic data points even for classes not present in the training set.

We introduce and investigate FAE (Foundation Auto-Encoders), a foundation generative-AI model for anomaly detection in time-series data, based on VAEs. FAE uses DC-VAE's network architecture \cite{dcvae_ieee_2023}, originally designed for multivariate anomaly detection. In particular, it leverages VAEs and Dilated Convolutional Neural Networks (DCNNs) to build a generic model for \emph{univariate time-series modeling}, which could eventually perform properly in out-of-the-box, zero-shot anomaly detection applications. The reasons for moving from multivariate to univariate time-series analysis are twofold: from an architectural point of view, we become independent of the spatial dimensionality of a MTS dataset -- i.e., we fix the spatial input dimensionality to one -- and can therefore apply exactly the same architecture without any modifications; from an analytics perspective, while a univariate model is at a disadvantage compared to a multivariate model -- i.e., it loses access to cross-correlational information, which we have shown might be critical for better data modeling \cite{wtmc2023, dcvae_ieee_2023} -- the univariate version puts the focus exclusively on the temporal behavior of the data, which is exactly the target of the generalization we are looking for -- we want a model that generalizes across the temporal dimension and not necessarily the spatial one, which varies among different problems. 

We introduce the main concepts and ideas behind FAE, along with its network architecture, and present preliminary results in the analysis of two different datasets: (1) a multi-dimensional network monitoring dataset -- TELCO \cite{skpg-0539-23}, collected from an operational mobile Internet Service Provider (ISP), recently released openly to the research community; and (2) a large-span dataset of 250 multi-domain time series released for the KDD2021 yearly data contest \cite{kdd2021}. The remainder of the paper is organized as follows: Section \ref{sec:related_work} presents an overview of the related work. In Section \ref{sec:fae} we describe the FAE model and its underlying network architecture. Section \ref{sec:experiments} reports the preliminary results obtained with FAE on the analysis of time-series from TELCO and KDD2021, with a particular focus on generalization and zero-shot modeling. Discussion on the potential of FAE, along with its limitations, is presented in Section \ref{sec:discussion}. Finally, Section \ref{sec:conclusions} concludes the paper.
 
\section{Related Work}\label{sec:related_work}

The diversity of data characteristics and types of anomalies results in a lack of universal anomaly detection models. Modern approaches to time-series anomaly detection based on deep learning technology have flourished in recent years \cite{Guansong2021, adts2022}. Due to their data-driven nature and achieved performance in multiple domains, generative models such as VAEs and Generative Adversarial Networks (GANs) \cite{goodfellowgan2014} have gained relevance in the anomaly detection field \cite{zavrak2020anomaly, zenati2018efficient, Chen2021AJM, donahue2016adversarial, li2019mad, geiger2020tadgan, ggg2021generative}.

VAEs \cite{kingma2013auto, doersch2016tutorial, kingma2019introduction} represent a powerful and widely-used class of models to learn complex data distributions. Unlike GANs, a potential limitation of VAEs is the prior assumption that latent sample representations are independent and identically distributed. While this is the most common assumption followed in the literature, there is ongoing research on the benefits of accounting for covariances between samples in time to improve model performance \cite{casale2018gaussian, girin2019notes, fortuin2020gp, ramchandran2021longitudinal}. For example, while the original work \cite{kingma2013auto} assumes that the prior over the parameters and latent variables are centered isotropic Gaussian and the true posteriors are approximately Gaussian with approximately diagonal covariance, \cite{fortuin2020gp} proposes an approximation capturing temporal correlations, by considering a Gaussian process prior in the latent space.

\begin{figure}[t!]
\centering
\includegraphics[width=\columnwidth]{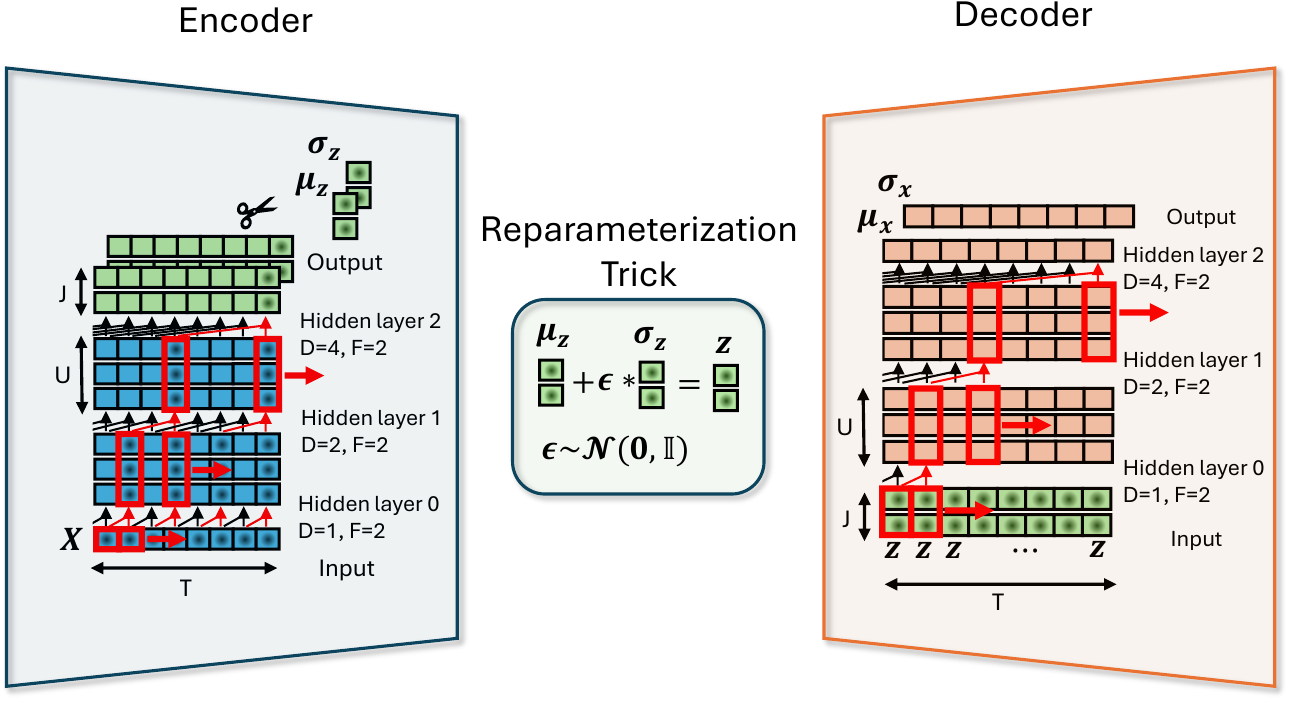}
\caption{FAE's encoder/decoder architecture using causal dilated convolutions, implemented through a stack of 1D convolutional layers.}
\label{fig_cd-conv}
\end{figure}

Modeling data sequences through a combination of variational inference and deep learning architectures has been vastly researched in other domains in recent years, mostly by extending VAEs to Recurrent Neural Networks (RNNs), with architectures such as STORN \cite{bayer2014learning}, VRNN \cite{chung2015recurrent}, and Bi-LSTM \cite{shabanian2017variational} among others. Convolutional layers with dilation have been also incorporated into some of these approaches \cite{yang2017improved, lai2018stochastic}, allowing to speed up the training process based on the possibilities of parallelization offered by these architectures. One of these approaches using Dilated Convolutional Neural Networks as the encoder-decoder architecture for VAEs is our DC-VAE model \cite{dcvae_ieee_2023, wtmc2022}.

Transformer-based models \cite{Vaswani2017} are gaining popularity in recent years for time-series analysis, given their remarkable performance in large-scale settings, such as long sequence time-series forecasting (LSTF). LSTF requires capturing long-range dependencies between input and output efficiently. Earlier examples include the TFT interpretable model \cite{LIM2021} and the MQTransformer model \cite{Chen2022}. The Informer model \cite{Zhou2021} introduced Transformers for long sequence forecasting through sparse self-attention mechanisms. This concept has since been further refined through various forms of inductive bias and attention mechanisms in models like the Autoformer \cite{autoformer2021} and the FEDformer \cite{fedformer2022}.

Finally, there is a recent surge in papers targeting the conception of foundation models for time-series data, capable of generating accurate predictions for diverse datasets not seen during training. The underlying concept of these models is to rely on highly expressive, large-scale architectures which are trained on millions or billions of time-series data points, coming from very diverse domains and having high heterogeneity in terms of temporal behaviors and characteristics. TimeGPT-1 \cite{garza2023timegpt1}, PromptCast \cite{promptcast2023}, LLMTime \cite{gruver2023large}, TimesFM \cite{das2024decoderonly}, Lag-Llama \cite{rasul2024lagllama}, and Time-LLM \cite{jin2023time} are all examples of novel foundation models for time-series forecasting, which target a ZSL application. FAE follows exactly this concept, but using a much smaller and simpler architecture. While this adds certain limitations in terms of expressiveness and therefore generalization capabilities, it also opens the door to the exploration of other venues, such as the combined utilization of smaller foundation models in the form of ensembles, in combination with domain detection strategies.

\section{FAE Model and Network Architecture}\label{sec:fae}

Time-series are generally processed through sliding windows, condensing the information of the most recent $T$ measurements. We define $\boldsymbol{X}$ as the input vector in $\mathbb{R}^{1 \times T}$. For a given input $\boldsymbol{X}$, the trained VAE model produces two different predictions, $\boldsymbol{\mu_X}$ and $\boldsymbol{\sigma_X}$ -- vectors in $\mathbb{R}^{1 \times T}$, corresponding to the parameterization of the probability distribution which better represents the given input. If the VAE model was trained (mainly) with data describing the normal behavior of the monitored system, then the output for a non-anomalous input would not deviate from the mean $\boldsymbol{\mu_X}$ more than a specific integer $\boldsymbol{\alpha}$ times the standard deviation $\boldsymbol{\sigma_X}$. On the contrary, if the input presents an anomaly, the output would not belong to this normality region. The main goal of the VAE model is to learn a compressed representation of $\boldsymbol{X}$ in an unsupervised manner. This compressed representation $\boldsymbol{Z}$ is referred to as a latent variable, and it is learned by training the VAE to generate data that is similar to the input data. VAEs learn a probabilistic mapping between the input data and its latent variable, which allows to generate new data by sampling from the learned latent variable distribution.

Figure \ref{fig_cd-conv} depicts the encoder/decoder architecture used in FAE, which is an adaptation of DC-VAE's architecture, for the case of univariate time-series analysis. The FAE model functions as a univariate model trained on various series within a system simultaneously, treating them as distinct classes of series. Similar to the original DC-VAE version, FAE allows for monitoring of all time-series within a MTS process using a single model, albeit analyzing one time-series at a time. The architecture, based on dilated convolutional neural networks (DCNNs), is capable to exploit the temporal dependence of values for longer sequences. The main difference with DC-VAE is that the new architecture has to accommodate univariate input samples $\boldsymbol{X} \in \mathbb{R}^{1 \times T}$, rather than multivariate ones. To maintain the concept of compression -- i.e., the dimension of the latent space $\boldsymbol{Z}$ has to be lower than the input dimension of $\boldsymbol{X}$ -- the latent space in FAE reduces dimensionality along the temporal dimension; in DC-VAE, the dimensionality reduction operates in the spatial dimension.

\begin{figure*}[t!]
\centering
\includegraphics[width=\textwidth]{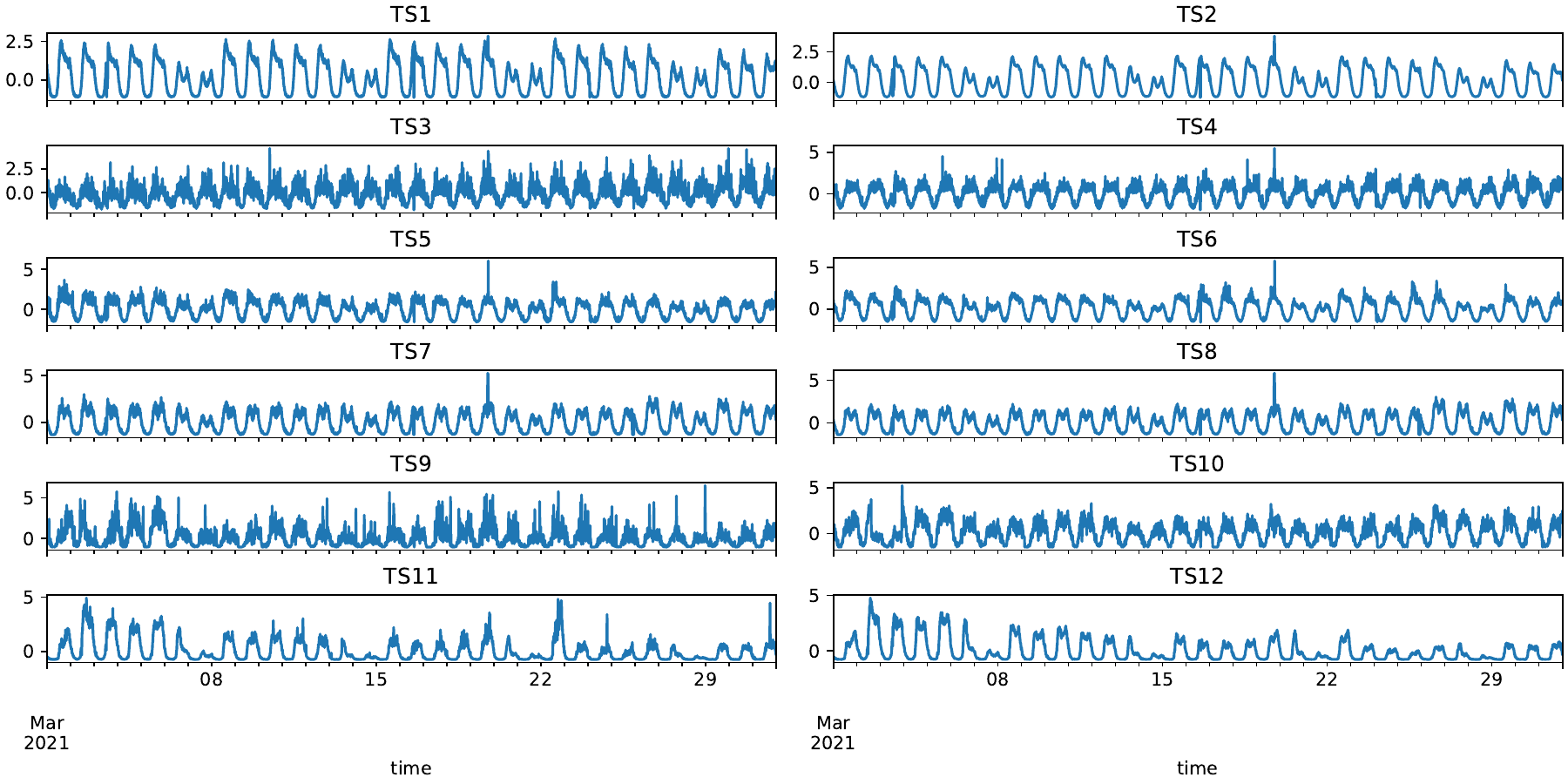}
\caption{TELCO time-series, for one month worth of data (March 2021), sampled at a five minutes rate.}
\label{fig:telco_example}
\end{figure*}

Using DCNNs forces to keep the sequence length $T$ at the output of each hidden layer -- referred to as $\boldsymbol{H}$ -- thus $\boldsymbol{H} \in \mathbb{R}^{U \times T}$, where $U$ is the number of filters in the layer. As shown in Figure \ref{fig_cd-conv}, the dilation of each layer at the encoder increases exponentially as the network deepens, ensuring that each time $t$ of the output at the final layer has information from all the previous times of $\boldsymbol{X}$ up to $t$, i.e., $[X_0, X_1, ... ,X_{t}]$. This means that the last sample of the output at position $T-1$ contains information from the entire input sequence $\boldsymbol{X}$. Then, two layers in parallel with $J$ filters of size one are applied at the output of the last hidden layer, bringing the output to a $J$-dimensional latent space. The output of the encoder results from keeping only the values at time $T-1$ at the output of these filters, resulting in vectors $\boldsymbol{\mu_Z}, \boldsymbol{\sigma_Z} \in \mathbb{R}^{J \times 1}$, which define the distribution in the latent space of the inputs $\boldsymbol{X}$, where $J<T$. Using the reparameterization trick \cite{kingma2013auto}, the latent vector $\boldsymbol{Z} \in \mathbb{R}^{J \times 1}$ is generated, corresponding to the encoding of observation $\boldsymbol{X}$, and is then fed into the decoder. The decoder remains the same as in DC-VAE, which is symmetric with respect to the encoder. However, given that its input requires a sequence of $T$ values and not a single one, the input to the decoder is generated by repeating the vector $\boldsymbol{Z}$ for a total of $T$ times, obtaining an input sample $\in \mathbb{R}^{J \times T}$. As a result, FAE's decoder extracts information from the same latent vector $\boldsymbol{Z}$ for each time $t$, to generate the output parameters $\boldsymbol{\mu_X}, \boldsymbol{\sigma_X} \in \mathbb{R}^{1 \times T}$, which are used to evaluate deviations from the input observation $\boldsymbol{X}$. If the FAE model was trained (mainly) with data describing the normal behavior of the analyzed time-series, then the value of a non-anomalous sample $\boldsymbol{X}_t$ at time $t$ would not deviate from the predicted mean $\boldsymbol{\mu_X}_t$ more than a specific integer $\alpha$ times the standard deviation $\boldsymbol{\sigma_X}_t$. On the contrary, if the sample is anomalous, it would not belong to the region determined by the predicted mean and standard deviation.

In terms of size of the architecture, an interesting characteristic of FAE is that its structure and number of layers is defined by the length $T$ of the sliding window. In particular, the number of hidden layers $N$ and the length of filters $F$ are related through the dilation factor $d=F^h$ of the DCNNs, which grows exponentially with the layer depth $n \in [0, N-1]$. Subsequently, $N$ is the minimum value that verifies: $T \leq 2*F^{N-1}$. In the architectural example (cf. Figure \ref{fig_cd-conv}), the window length is $T=8$ and the filter length is $F=2$, and the target is achieved by taking $N=3$ hidden layers. This direct relationship between $T$ and the network architecture has a strong practical impact, making it easy to construct the encoder/decoder based on the desired temporal-depth of the analysis. As a final reference of architectural complexity, for a relatively small FAE architecture, using $T=256$ samples (less than one day of samples, at a 5' sampling-rate), the network exposes roughly half a million free parameters to train. Training FAE with a sufficiently large and heterogeneous training set, comprising multiple time-series of different characteristics, enhances its capability to generalize to unseen data and eventually to different domains.

\section{FAE Time-Series Prediction in the Practice}\label{sec:experiments}

\subsection{Application of FAE in TELCO}

We experiment with FAE in the analysis of an open MTS dataset arising from the monitoring of an operational mobile ISP, consisting of time-series with different structural properties. Referred to as the TELCO dataset~\cite{skpg-0539-23}, this \emph{large-scale} -- about 750 thousand samples, \emph{long time-span} -- seven months' worth of measurements (January 1st to July 31st, 2021) collected at a five-minutes scale, \emph{multi-dimensional} -- twelve different time-series, network monitoring dataset includes ground-truth labels for anomalous events at each individual time-series, manually labeled by the experts of the network operation center (NOC) managing the mobile ISP. The twelve time-series are typical data monitored in a mobile ISP, including volume of data traffic, number of SMS, number and amount of prepaid data transfer fees, number and cost of calls, etc. 

In this paper we focus on a more qualitative analysis of FAE's performance, focusing on its ability to properly track and reconstruct the different TELCO time-series, and the KDD2021 time-series in section \ref{sec:kdd2021}. Figure \ref{fig:telco_example} depicts a one-month example from the complete TELCO MTS dataset. Different time-series expose different behaviors, e.g., some of them are noisier (TS$_3$ and TS$_9$), others have lower dynamic ranges (TS$_{1}$), and some others show a smoother evolution (TS$_{2}$). All time-series exhibit daily seasonality, but some behave differently on weekends compared to workdays, while others show monthly trends either ascending or descending.

\begin{table}[t!]
\caption{Grid of hyperparameters used in the model calibration.}
\centering
\footnotesize
\renewcommand{\arraystretch}{1.2}
\begin{tabular}{|l||l|c|}
\hline
Hyperparameter & Grid Search Ranges & Best\\\hline\hline
$T$ - sequence length & $\lbrace 128-512 \rbrace$, step=32 & 256\\\hline
$J$ - latent dimension & $\lbrace 16-T/4 \rbrace$, step=16 & 48\\\hline
$\gamma$ - learning rate & $\lbrace 1e^{-5}-5e^{-4} \rbrace$ & $6e^{-5}$\\\hline
$m$ - mini-batch size & $\lbrace 16-96 \rbrace$, step=16 & 32\\\hline
$U$ - number of filters & $\lbrace 16-128\rbrace$, step=16 & $128$\\\hline
\end{tabular}
\label{tab:hyperparameter}
\end{table}

\subsection{Hyperparameter Search and Training}

One of the most important aspects when working with deep learning models is the search of model and training hyperparameters, along with the subsequent training of the model. Table \ref{tab:hyperparameter} shows the grid used for the hyperparameter search, as well as the best values (smallest validation loss), identified by Tree-structured Parzen Estimator (TPE) search \cite{bergstra2011algorithms}. In total, 50 attempts were tested on the grid. In the table, $T$ corresponds to the sequence length, and $J$ is the dimensionality of the latent space. Training hyperparameters include the learning rate $\gamma$ and the mini-batch size $m$. Finally, $U$ is the number of filters for each hidden convolutional layer, which together with the number of layers and the input and output dimensions define the size of the architecture in terms of the number of trainable parameters $p$. Considering the five minutes sampling rate of the time-series, the selected sequence length of $T = 256$ samples corresponds to a time window of 21hs and 20 minutes. The exact number of trainable parameters in the identified architecture is $p = 483.840$. 

We split the full, 7-months dataset in three independent, time-ordered sub-sets, using measurements from January to March for model training (3 months), April for model validation (1 month), and May to July for testing purposes (3 months). One of the disadvantages of the FAE model as compared to the multivariate DC-VAE model is the training time, and hence the time required for hyperparameter search. The FAE model requires approximately five times more training time than DC-VAE.

\begin{figure}[t!]
\centering
$\begin{array}{cc}
\hspace{-4mm}\includegraphics[width=0.52\columnwidth]{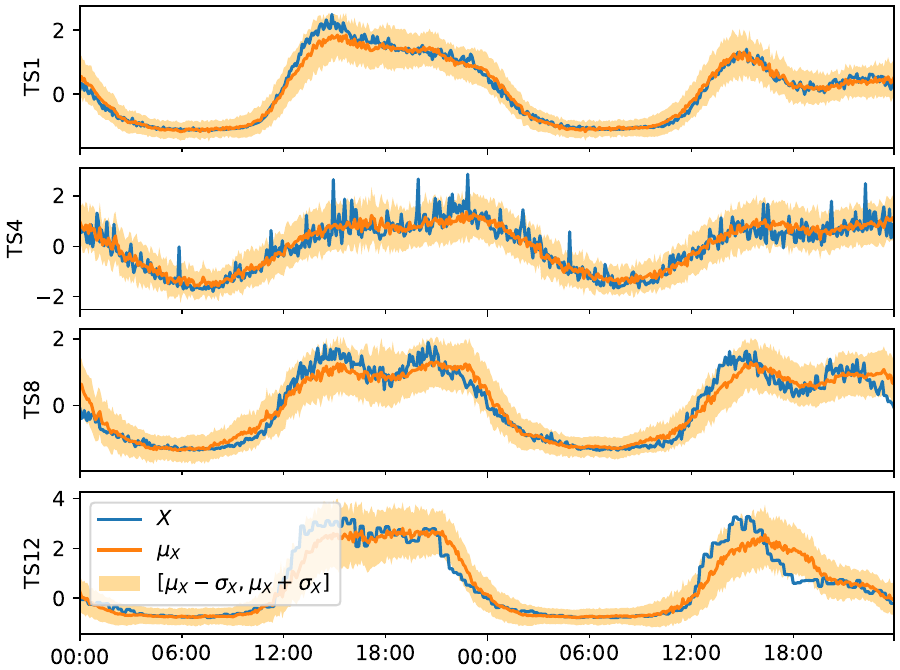} & \hspace{-2mm}\includegraphics[width=0.48\columnwidth]{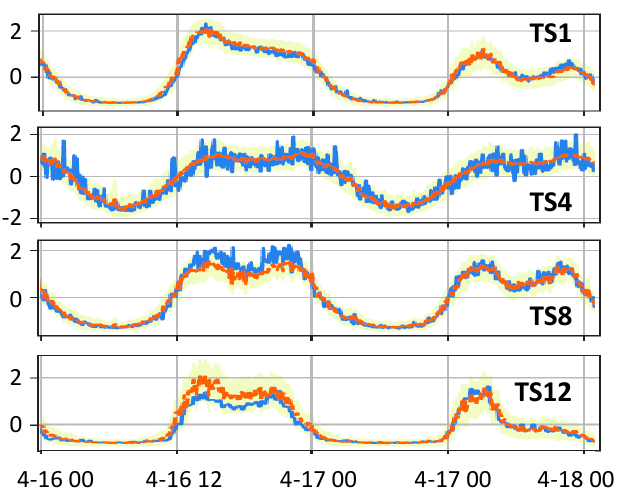}\\ 
\hspace{-4mm}\text{\scriptsize(a) Fully-trained FAE.} & \hspace{-2mm}\text{\scriptsize(b) DC-VAE Multivariate.}\\
\end{array}$
\caption{Predictions with fully-trained FAE and multivariate DC-VAE in two days of TELCO samples.}
\label{fig:prediction_example}
\end{figure}

\begin{figure}[t!]
\centering
\includegraphics[width=0.8\columnwidth]{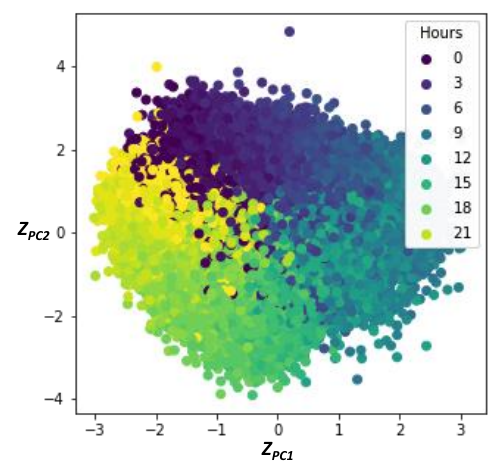}
\caption{Latent space representation -- temporal evolution.}
\label{fig:latent_space_hours}
\end{figure}

\begin{figure*}[t!]
\renewcommand{\arraystretch}{1}
\centering
$\begin{array}{ccc}
\includegraphics[width=0.3\textwidth]{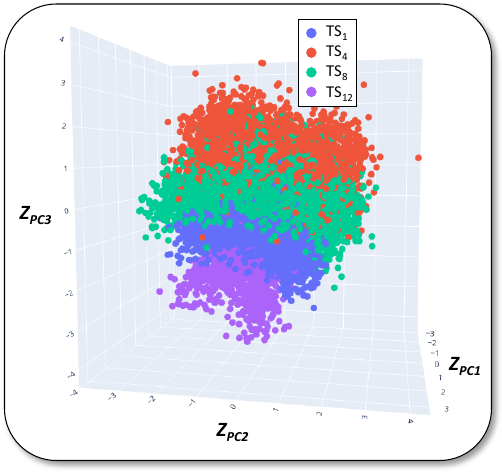} & \includegraphics[width=0.3\textwidth]{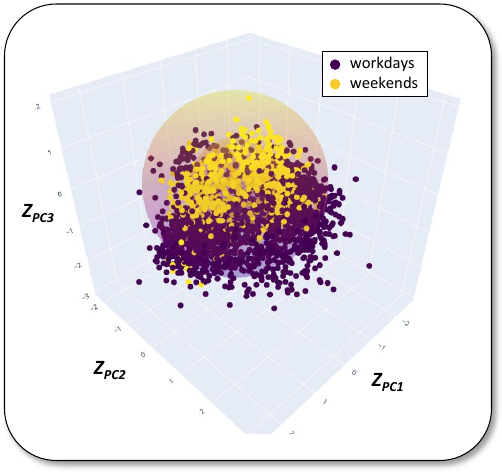} & \includegraphics[width=0.3\textwidth]{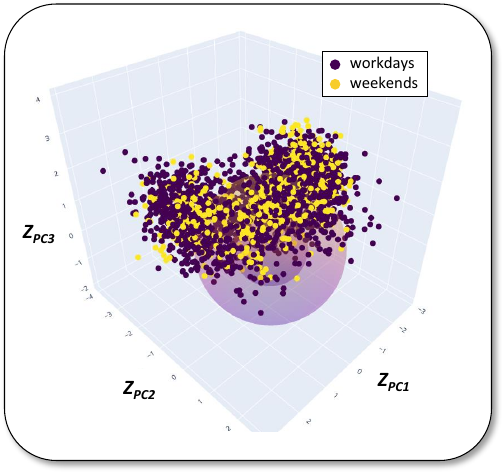}\\\vspace{-0.2mm}
\text{\footnotesize(a) Latent representations per time-series.} & \text{\footnotesize(b) Workdays vs weekends, TS$_{1}$.} & \text{\footnotesize(c) Workdays vs weekends, TS$_{4}$.}\\
\end{array}$
\caption{Latent space representation, (a) per different time-series (TS$_{1}$, TS$_{4}$, TS$_{8}$, TS$_{12}$), and (b,c) specifically for TS$_{1}$ and TS$_{4}$ in a temporal basis, considering workdays (purple) and weekends (yellow).}
\label{fig:latent_space_series_weekdays}
\end{figure*}

\begin{figure}[t!]
\centering
\includegraphics[width=0.75\columnwidth]{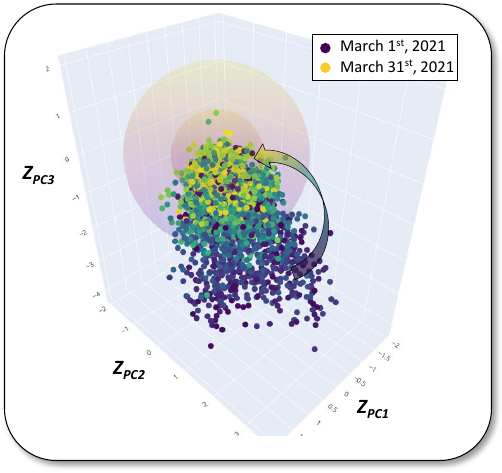}
\caption{Latent space representation for TS$_{12}$, in a daily basis -- from day 1 in purple to day 31 in yellow, for a full month.}
\label{fig:TS12_ls_days}
\end{figure}

\subsection{FAE Modeling Performance}

We evaluate the prediction performance of FAE in samples from the testing set, considering training on the full three months of data, for the 12 time-series, i.e., more than 300.000 samples. Figure \ref{fig:prediction_example}(a) depicts the resulting predictions $\boldsymbol{\mu_X}_t$ and $\boldsymbol{\sigma_X}_t$ for two days of testing samples $\boldsymbol{X}_t$ from June 2021, for four representative time-series, including TS$_{1}$, TS$_{4}$, TS$_{8}$, and TS$_{12}$. To add more variability, we consider a working day (Friday 4th) and a weekend day (Saturday 5th). FAE can properly track different types of behavior in the time-series, including the strong seasonal daily component, but also the operation during workdays and weekends, clearly visible in TS$_{12}$. Interesting to note is how different periods of time-series variability result in more or less tight normal-operation regions estimated by FAE, as defined by $\boldsymbol{\sigma_X}_t$. Figure \ref{fig:prediction_example}(b) depicts the predictions obtained by the former DC-VAE multivariate model in the same four time-series, using a different time period in April -- in this case from the validation test -- from Friday 16th till Saturday 17th. Results are similar, but in particular for TS$_{12}$, DC-VAE can better capture the drop observed on Saturday evening, exploiting the strong spatial correlation observed on Saturday between TS$_{12}$, TS$_{11}$, and both TS$_{1}$ and TS$_{2}$ (cf. Figure \ref{fig:telco_example}). Nevertheless, note that FAE predictions are slightly better than DC-VAE's for TS$_{8}$ and TS$_{12}$ on Friday.

To better understand the modeling capabilities of FAE, we focus on the analysis of the latent space $\boldsymbol{Z}$, for the different time-series and the different times of the analysis. Recall that the latent space set for FAE in this analysis is $J = 48$; to easily visualize $\boldsymbol{Z}$ as a two- or three-dimensional space, we apply standard PCA analysis, and study the top-two and top-three principal components $\boldsymbol{Z_{PCi,\dots i=1,2,3}}$. Figure \ref{fig:latent_space_hours} shows the latent representation of each sample $\boldsymbol{X}_t$ for a single day, for all the 12 time-series, depicting the encoded samples in colors, each color representing a different three-hour period of the day. Plotting the first two principal components shows that each hour period maps to a certain position in the latent space, and interestingly, the direction mirrors the progression of hours on a clock, ordered continuously by hour of the day.

Figure \ref{fig:latent_space_series_weekdays}(a) depicts now the first three principal components, displaying the previously analyzed four time-series TS$_{1}$, TS$_{4}$, TS$_{8}$, and TS$_{12}$, independently. FAE maps each time-series to a different region of the latent space, showing it can properly differentiate among different time-series characteristics. Closely located samples from different time-series exhibit similar behaviors in the time-series space. For example, time-series TS$_{1}$ and TS$_{12}$ are closer to the center of the latent space, and both exhibit a similar behavior (cf. Figure \ref{fig:telco_example}), with marked differences between weekends and workdays. On the other hand, time-series TS$_{4}$ and TS$_{8}$ have a similar temporal behavior without marked variations between workdays and weekends, and are located together and farther from the center of $\boldsymbol{Z}$.

The difference between time-series for workdays and weekends is further explored in Figures \ref{fig:latent_space_series_weekdays}(b,c), where workday samples are displayed in purple color, and weekends in yellow, for (b) time-series TS$_{1}$ and (c) time-series TS$_{4}$. The plots include two spheres with radius one and two, reflecting the expected Gaussian distribution of the latent space. The difference between workdays and weekends are clear for TS$_{1}$, with weekends clustering closer to the center (smaller dynamic range, cf. Figure \ref{fig:telco_example}), and workdays located closer at the sphere borders (bigger dynamic range). This workdays/weekends difference is not observed for TS$_{4}$. Note in Figure \ref{fig:telco_example} how time-series TS$_{11}$ and TS$_{12}$ exhibit a downtrend behavior during the month, which can also be observed in the latent-space. Figure \ref{fig:TS12_ls_days} depicts the latent representation of time-series TS$_{12}$, where days of the month are differentiated by color, from day 1 in purple to day 31 in yellow. As the month goes by, the representation in the latent space moves from the outside borders closer towards the center.

\begin{figure}[t!]
\centering
\includegraphics[width=\columnwidth]{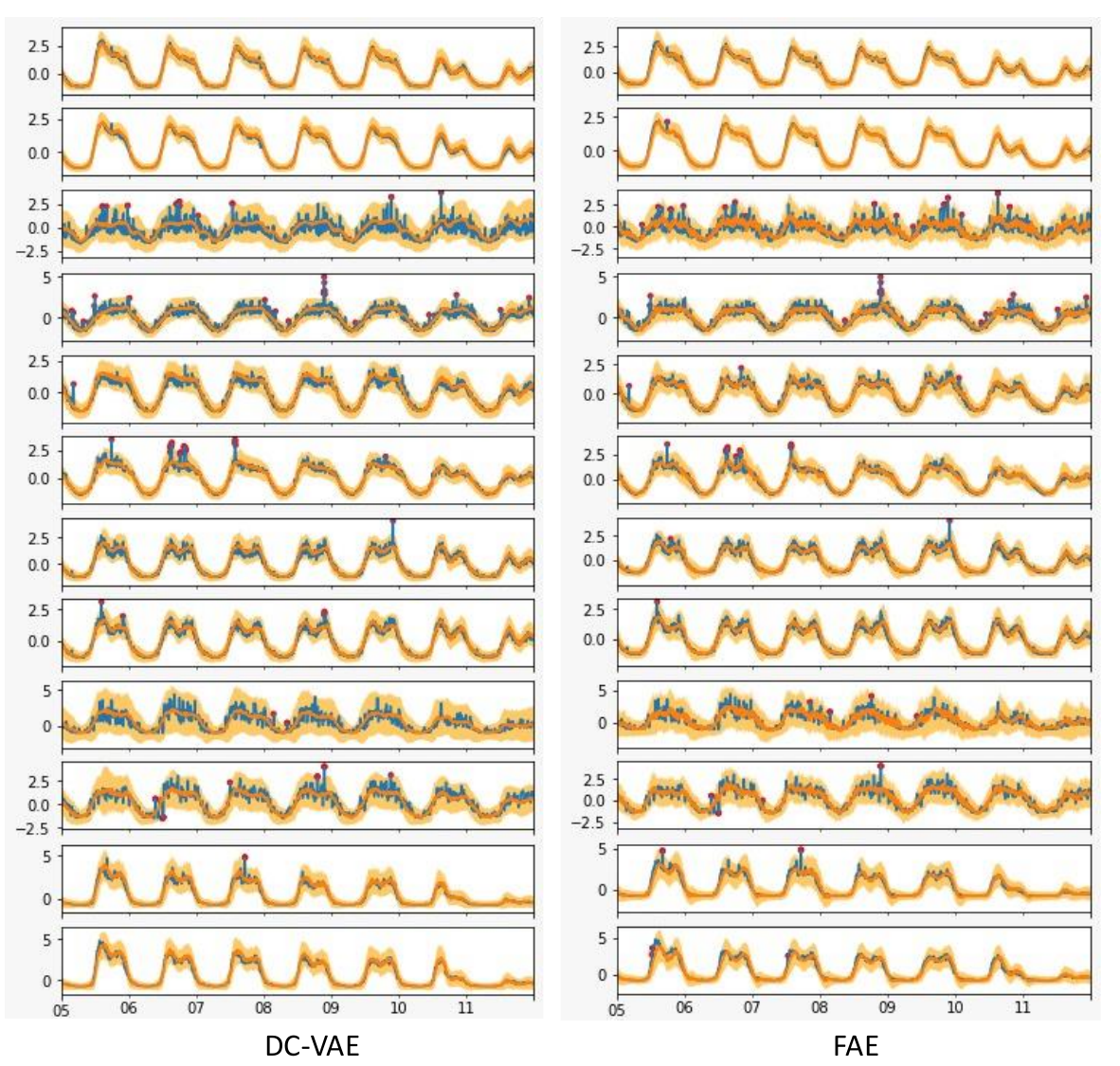}
\caption{FAE vs DC-VAE in TELCO.}
\label{fig:telco}
\end{figure}

Finally, Figure \ref{fig:telco} reports the prediction results, along with flagged anomalies, for both DC-VAE (left) and FAE (right) for the complete 12 time-series in TELCO. Modeling behavior and detection performance are comparable between both models.

To wrap-up these preliminary evaluations, we observe how FAE can properly capture and differentiate among the different temporal behaviors present in the time-series used for training, suggesting a sufficiently expressive model and architecture to model a large and heterogeneous dataset of time-series. In addition, the visual analysis of the latent representations in FAE evidences how VAEs -- despite their generative nature -- are rather transparent in their operation and behavior, making interpretation and analysis simpler and more human-friendly. This is indeed a strong advantage of VAEs as a powerful yet explainable generative AI model, as compared to modern generative AI approaches, which operate in a more black-box manner.

\begin{figure}[t!]
\centering
\includegraphics[width=\columnwidth]{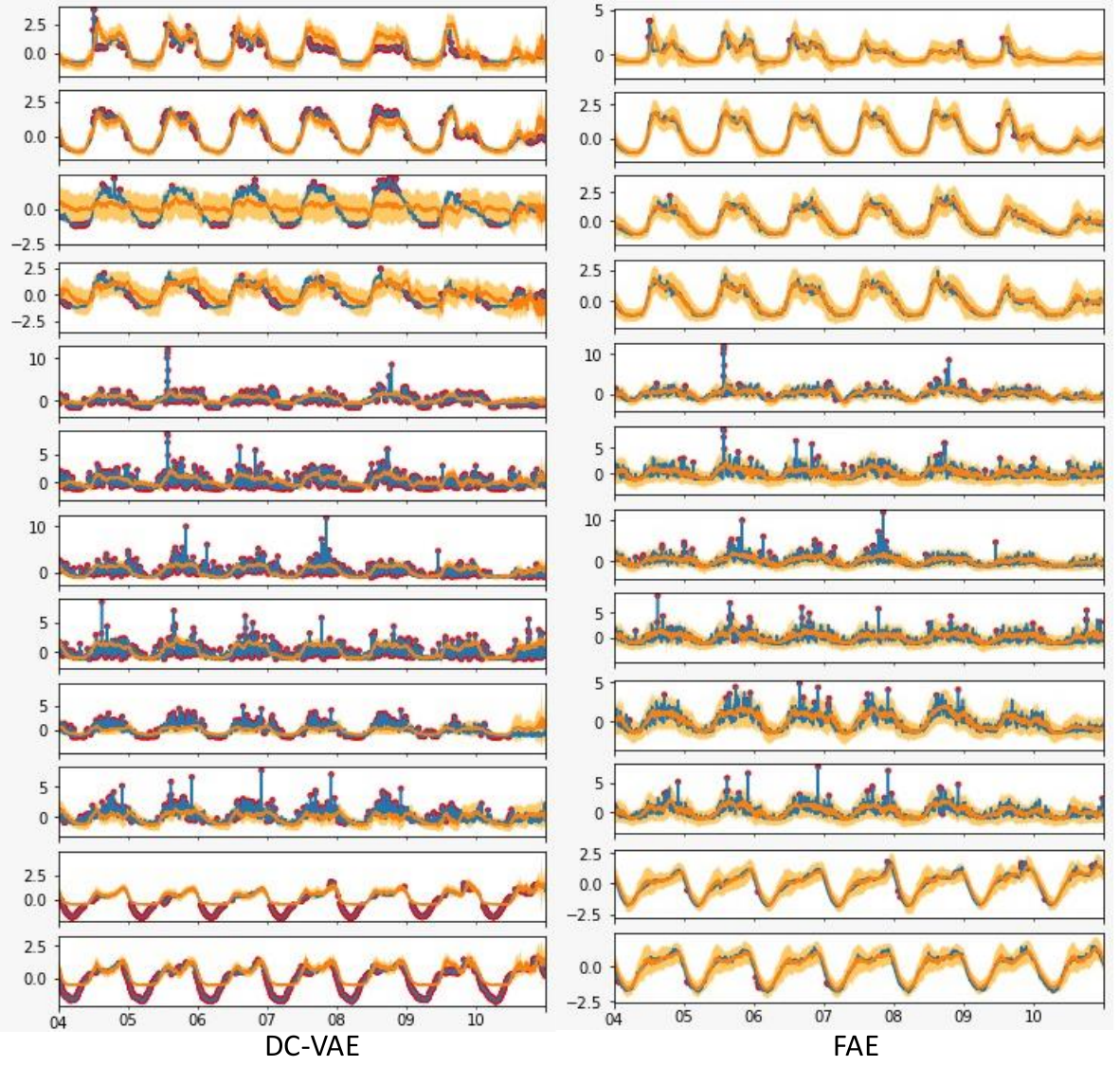}
\caption{FAE vs DC-VAE in TELCO2.}
\label{fig:telco2}
\end{figure}

\subsection{FAE vs DC-VAE in TELCO2}

What happens when we modify the application domain? In Figure \ref{fig:telco2}, we apply the pretrained DC-VAE and FAE models in a modified version of TELCO, corresponding to newer data captured in 2024, with modifications in some of the time-series. We refer to this new dataset as TELCO2. On the left side, we depict the modeling and detection performance of DC-VAE, and FAE on the rightside. Both models are pretrained in TELCO, and then applied on TELCO2. In this scenario, it is clear to see the flexibility and advantages of FAE over a multivariate version of the model, which would need re-calibration to cope with the slight yet present variations in the domain of the dataset. In particular, the spatial modification of correlations and time-series' characteristics make it challenging for DC-VAE to properly adjust to this domain modification.

\subsection{Zero-shot Modeling Behavior}

We investigate now the performance of FAE in a zero-shot setting, testing the model for time-series not seen at training time. We focus the analysis on TS$_{12}$, due to its combined seasonality and particular temporal trend, as well as its strong correlated behavior to TS$_{11}$ (cf. Figure \ref{fig:telco_example}). We train FAE on three different training datasets from TELCO, using the same 3/1/3-months temporal split for training/validation/testing, but considering a different number of time-series TS$_{i}$. The first model uses all 12 time-series -- we refer to it as \emph{full-FAE}; the second model considers a zero-shot setting for TS$_{12}$, with a training dataset which includes time-series TS$_{1}$ to TS$_{11}$, leaving out all samples from TS$_{12}$; given the strong temporal correlation between TS$_{12}$ to TS$_{11}$, we also train a third model leaving out all samples from TS$_{11}$ and TS$_{12}$, i.e., training on time-series TS$_{1}$ to TS$_{10}$.

The full FAE model mimics a situation where we pretrain with a sufficiently large and heterogeneous dataset which covers the statistical behavior of the downstream data -- i.e., a model that \emph{has seen it all}. The other two models mimic two different levels of zero-shot learning: the former represents a pure zero-shot setting for TS$_{12}$, where the pretrained model has nevertheless observed a similar statistical behavior in a different time-series, i.e., TS$_{11}$ -- in particular, it has seen both the seasonality and the monthly trend behaviors; the latter represents a more challenging setting, where the pretrained model has not seen the monthly trend behavior, which is not present in TS$_{1}$ to TS$_{10}$.

\begin{figure}[t!]
\renewcommand{\arraystretch}{1}
\centering
$\begin{array}{c}
\hspace{-4mm}\includegraphics[width=\columnwidth]{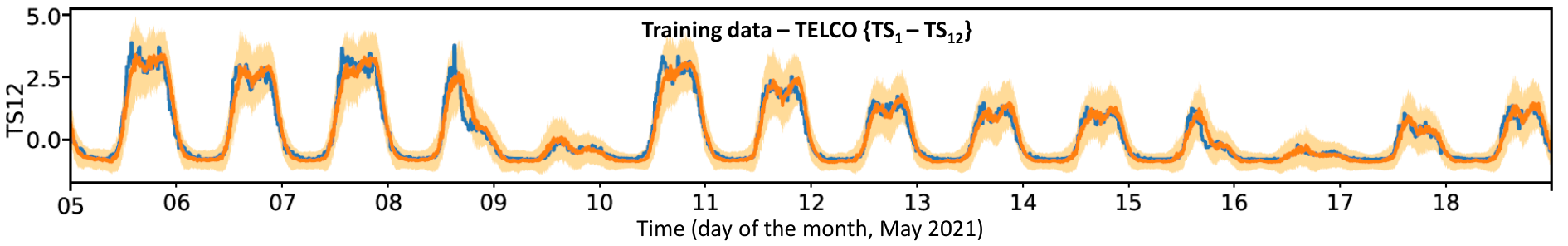}\\
\text{\footnotesize(a) FAE predictions for TS$_{12}$, with full-FAE (12 time-series).}\\~\\
\hspace{-4mm}\includegraphics[width=\columnwidth]{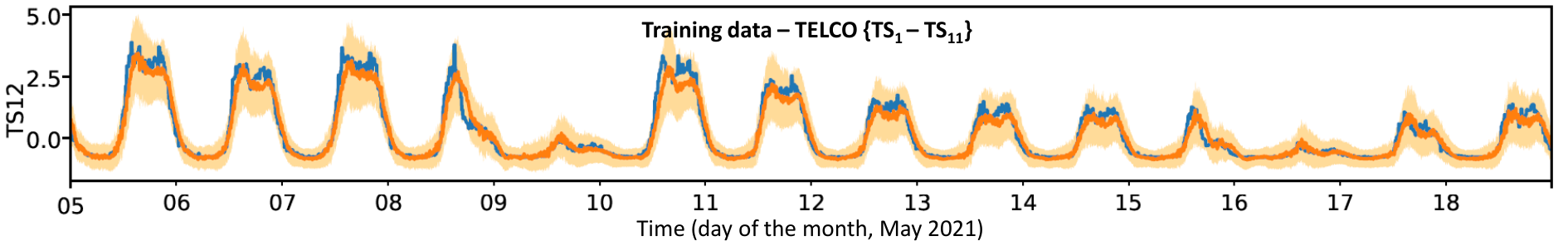}\\
\text{\footnotesize(b) FAE predictions for TS$_{12}$, with FAE trained without TS$_{12}$.}\\~\\
\hspace{-4mm}\includegraphics[width=\columnwidth]{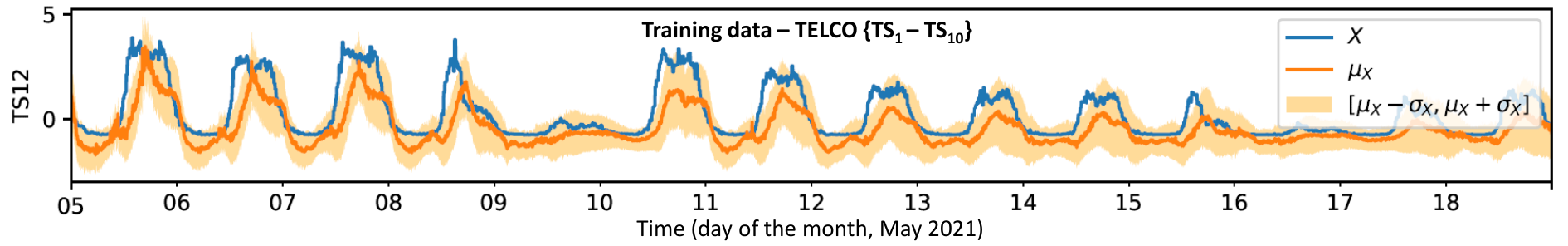}\\
\text{\footnotesize(c) FAE predictions for TS$_{12}$, with FAE trained without TS$_{11}$ and TS$_{12}$.}\\
\end{array}$
\caption{Zero-shot modeling experimentation, predicting TS$_{12}$ for two weeks in the testing dataset (May 2021). (a) FAE is trained on the full, 12 time-series training set -- modeling performance is optimal. (b) FAE is trained on 11 time-series, leaving out TS$_{12}$ -- performance remains almost unchanged. (c) FAE is trained on 10 time-series, leaving out TS$_{11}$ and TS$_{12}$ -- modeling performance is impacted.}
\label{fig:predictions_exp_drop}
\end{figure}

Figure \ref{fig:predictions_exp_drop} presents the prediction performance of the three models, when applied to two weeks of TS$_{12}$ samples, from May 5 to May 19, 2021. In Figure \ref{fig:predictions_exp_drop}(a), the modeling performance for full-FAE is optimal, as it can properly track the different behaviors and patterns in the time-series, similarly to Figure \ref{fig:prediction_example}(a). A similar performance is observed in Figure \ref{fig:predictions_exp_drop}(b) for the second model, which learns the characteristics of TS$_{12}$ at training time, from TS$_{11}$. Not surprisingly, the performance of the third model in Figure \ref{fig:predictions_exp_drop}(c) is significantly worse than for the other two models, given the lack of a similar temporal pattern in the training data. To some extent, there is an identification with the patterns observed in time-series TS$_{1}$ -- note how the daily sharp peaks are exacerbated -- which is coherent with their close representations in the latent space (cf. Figure \ref{fig:latent_space_series_weekdays}(a)). Nevertheless, it somehow manages to capture and track the monthly downtrend, even without previous evidence of it.

\subsection{Application of FAE in KDD2021}\label{sec:kdd2021}

In this last evaluation, we extend the application of FAE to the KDD2021 dataset, consisting of 250 different time-series arising from many different domains. The dataset is part of the larger UCR Time Series Classification Archive (\url{https://www.cs.ucr.edu/~eamonn/time_series_data_2018/}). Within the dataset, each time-series comes in two parts, a train partition and a test partition, which are generally of different sizes. For consistency with TELCO, we take 12 different time series from KDD2021 and apply FAE on them, training on the train partition and reporting testing results. We refer to this dataset as KDD2021-12.

\begin{figure}[t!]
\centering
\includegraphics[width=\columnwidth]{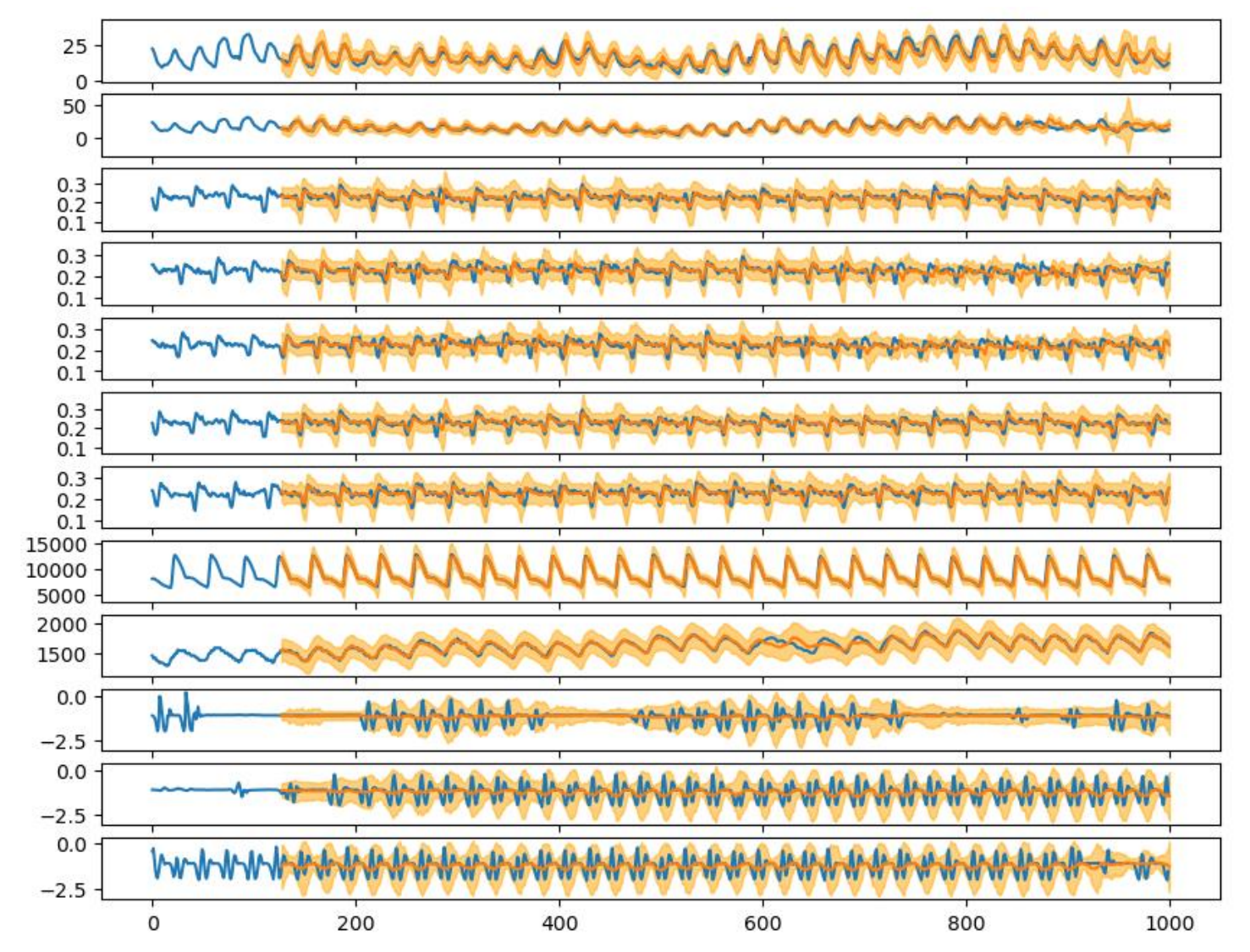}
\caption{FAE applied in 12 time-series from KDD2021.}
\label{fig:kdd2021}
\end{figure}

\begin{figure}[t!]
\centering
\includegraphics[width=\columnwidth]{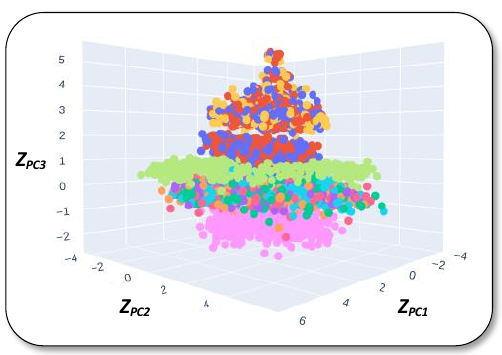}
\caption{FAE's latent space for the 12 KDD2021 time-series.}
\label{fig:kdd2021_latent}
\end{figure}

Figure \ref{fig:kdd2021} depicts the predictions obtained with FAE in KDD2021-12. Tracking performance is highly accurate, in particular when considering the anomaly detection application which FAE targets, where the main goal is not to fully reconstruct the output, but rather to provide a mean and variance values where samples are expected to be contained. A couple of observations are worth considering: first, there is a much higher heterogeneity in KDD2021-12 as compared to TELCO, yet FAE manages to properly track the expected normal behavior of the time-series. Second, for some of the KDD2021-12 time series -- e.g., TS$_{10}$, there are long periods of inactivity (constant value) followed by a very spiky behavior; in those cases, note how FAE provides a proper prediction contained in the forecasted variance, which covers the actual time-series values. Finally, Figure \ref{fig:kdd2021_latent} depicts the resulting latent space for the KDD2021-12 time-series, which evidences a more complex mix to model, evidenced by clouds of dots from different time-series mixed together; nevertheless, there is a clear structure in the data -- clusters with single time-series or with time-series with similar behavior, which again confirms the principled concepts employed by FAE when it comes to the application of VAEs to understand the modeling approach.

\section{Discussion and Limitations of FAE}\label{sec:discussion}

Selecting VAEs for our foundation model exploration has both benefits and limitations, which we briefly discuss next. On the benefits side, we have shown how easy it is to explore and interpret the functioning of the encoding and the behavior of the encoded time-series in the latent space, making the model transparent and easy to tame, particularly for training. While VAEs may struggle with capturing long-range dependencies in the data, we have shown how the integration of DCNNs as part of the encoding/decoding networks enables tracking multiple different temporal behaviors in the time-series, from seasonality to long-term trends.

FAE shows potential to be a strong foundation model for time-series analysis, but so far, we have only trained and tested the model with a rather small number of datasets from a limited number of different domains, and thus still require further assessment. Indeed, thorough testing and validation would be necessary to assess FAE's performance and generalization capabilities in different scenarios. Ultimately, the effectiveness of FAE as a foundation model would depend on its performance in various real-world scenarios and its ability to generalize to different datasets and different domains. 

While powerful, VAEs may have limitations in terms of expressiveness when tasked with learning and mapping a large-scale number of highly heterogeneous time-series data. VAEs operate under the assumption of a latent variable space with a simple distribution, which may not always capture the intricate and diverse characteristics of more complex and diverse time-series data.

We note that the performance of FAE in generalization and zero-shot learning tasks can be affected by factors such as the complexity of the data, the dimensionality of the latent space, the choice of the encoder/decoder architecture, and the quality and diversity of the training data. Additionally, while VAEs can capture global structure in the data distribution, they may not always capture fine-grained details or handle complex data distributions as effectively as other generative models. Finally, in terms of scalability, FAE's performance may degrade with extremely large datasets or highly heterogeneous data, as the model complexity may need to increase significantly.

\section{Concluding Remarks}\label{sec:conclusions}

We have introduced FAE, a novel approach for time-series modeling, motivated by the performance realized by large pretrained foundation models in different domains. FAE targets the detection of anomalies in univariate time-series data, leveraging VAEs and DCNNs to pretrain on large-scale, heterogeneous time-series datasets, potentially enabling to properly model and track a baseline for normal operation, even on unseen datasets.

The preliminary assessment of FAE's performance has shown promising results. In particular, we have provided evidence of FAE's capabilities to capture and distinguish various temporal behaviors within the training time-series, demonstrating a promising capacity to model large and heterogeneous datasets effectively. The interpretability of FAE's latent representations showcased VAEs' transparency in operation, facilitating simpler analysis and interpretation compared to black-box generative AI models.

Our exploration extended to the zero-shot learning scenario, where FAE's performance on unseen time-series was assessed. We tested FAE in three settings, ranging from optimal modeling performance to more challenging scenarios. While FAE performs properly in tracking different behaviors and patterns in the time-series, even in the absence of previous evidence, there is room for further improvement, especially in capturing previously unseen temporal trends.

These initial findings underscore FAE's potential as a feasible foundation model for time-series analysis. However, it is essential to note that our evaluation was limited to a rather small number of datasets from a limited number of different domains. As part of our ongoing work, we are focusing on comprehensively testing FAE's performance and generalization capabilities across diverse scenarios, using much larger and heterogeneous time-series datasets for training, considering more heterogeneous domains.

\section*{Acknowledgment}

This work has been partially supported by the FWF Austrian Science Fund, Project I-6653 \emph{GRAPHS4SEC}, the Austrian FFG ICT-of-the-Future project \emph{DynAISEC -- Adaptive AI/ML for Dynamic Cybersecurity Systems} -- ID 887504, and by the Uruguayan CSIC project with reference CSIC-I+D-22520220100371UD \emph{Generalization and Domain Adaptation in Time-Series Anomaly Detection}.

\balance
\bibliographystyle{ACM-Reference-Format}
\bibliography{biblio}

\end{document}